\documentclass[journal,twoside,web]{ieeecolor}
\usepackage{generic}
\usepackage{cite}
\usepackage{amsmath,amssymb,amsfonts}
\usepackage{algorithmic}
\usepackage{graphicx}
\usepackage{algorithm,algorithmic}
\usepackage{hyperref}
\hypersetup{hidelinks}
\usepackage{textcomp}
\usepackage{fontawesome5}
\usepackage{caption}

\usepackage{subcaption}

\usepackage{booktabs}
\usepackage{multirow}
\usepackage{pifont}

\newcommand{\Tool}[0]{GlucoTune}

\def\BibTeX{{\rm B\kern-.05em{\sc i\kern-.025em b}\kern-.08em
    T\kern-.1667em\lower.7ex\hbox{E}\kern-.125emX}}
\begin{document}
\title{\Tool{}: A Unified Framework for Blood Glucose Preprocessing, Forecasting, and Benchmarking in Diabetes}

\author{Davide Marelli, Giorgia Rigamonti, Mirko Paolo Barbato, Paolo Napoletano,~\IEEEmembership{Senior~Member,~IEEE}%
\thanks{The authors are with the Department of Informatics, Systems and Communication, University of Milano-Bicocca, Milano 20126, Italy (e-mail: davide.marelli@unimib.it; giorgia.rigamonti@unimib.it; mirko.barbato@unimib.it; paolo.napoletano@unimib.it). Correspondin author: Davide Marelli. (Davide Marelli and Giorgia Rigamonti contributed equally to this work.)}
}

\maketitle

\begin{abstract}  %% MAX 250 words

Preprocessing blood glucose time-series data is a critical yet often overlooked step in developing data-driven methods for diabetes management, particularly for type 1 diabetes. The lack of standardized preprocessing workflows and evaluation protocols hinders reproducibility and complicates fair comparison across studies. These challenges are further exacerbated by data-sharing restrictions, as privacy and licensing constraints often prevent the redistribution of preprocessed medical datasets.
To address these limitations, we present \Tool{}, a comprehensive and extensible framework for reproducible experimentation with blood glucose time-series data. The framework standardizes the entire experimental workflow, from preprocessing to model evaluation, enabling reproducible experiments directly from the original datasets. Reproducible preprocessing is achieved through configurable pipelines defined in portable YAML configuration files, ensuring consistent data handling without distributing sensitive preprocessed data.
Beyond preprocessing, \Tool{} provides a unified interface for implementing, training, and evaluating blood glucose prediction models. The framework integrates public datasets through standardized wrappers and provides a curated collection of state-of-the-art blood glucose prediction and general time-series forecasting methods, while remaining readily extensible to additional datasets, preprocessing strategies, and forecasting models.
To promote transparent and consistent evaluation, \Tool{} includes a benchmarking leaderboard that reports results across datasets, preprocessing configurations, and forecasting methods, enabling systematic comparison of experimental settings. A graphical user interface (GUI) further improves accessibility by allowing users to configure, execute, and compare experiments with minimal programming effort. We demonstrate the effectiveness of \Tool{} through comprehensive experiments and assess its usability in a user study.

\end{abstract}


\section{Introduction}
\label{sec:introduction}
% \IEEEPARstart{D}{iabetes} mellitus is a chronic condition characterized by insufficient or absent insulin production, leading to imbalances in Blood Glucose Concentration (BGC)~\cite{american2010diagnosis}. Persistent hyperglycemia (BGC $>$ 180 mg/dL) increases the risk of vascular complications~\cite{hyper}, while hypoglycemia (BGC $<$ 70 mg/dL) can cause severe adverse outcomes~\cite{hypoglycemia}. 
\IEEEPARstart{D}{iabetes} mellitus is a chronic condition characterized by insufficient or absent insulin production, leading to impaired regulation of Blood Glucose Concentration (BGC)~\cite{american2010diagnosis}. Persistent hyperglycemia (BGC $>$ 180 mg/dL) is associated with an increased risk of long-term vascular complications~\cite{hyper}, whereas hypoglycemia (BGC $<$ 70 mg/dL) may result in severe acute adverse events~\cite{hypoglycemia}. 

%Continuous glucose monitoring (CGM) has revolutionized diabetes management by enabling real-time intervention and facilitating predictive modeling of glucose fluctuations~\cite{cgm}. 
% With the growing availability of Continuous Glucose Monitoring (CGM) data, Machine Learning (ML) and Deep Learning (DL) techniques have become central to BGC prediction. A wide range of models has been explored, including ARIMA~\cite{arima}, Random Forests~\cite{blood}, CNNs~\cite{freiburghaus2020deep}, RNNs~\cite{rnn,convolutional}, and Transformers~\cite{xue2024bgformer,bian2024hybrid}.
The widespread adoption of Continuous Glucose Monitoring (CGM) devices has enabled the development of a broad range of Machine Learning (ML) and Deep Learning (DL) approaches for BGC prediction. Existing methods span classical statistical models, such as ARIMA~\cite{arima}, traditional machine learning approaches, such as Random Forests~\cite{blood}, and modern deep learning architectures including CNNs~\cite{freiburghaus2020deep}, RNNs~\cite{rnn,convolutional}, and Transformers~\cite{xue2024bgformer,bian2024hybrid}.

As the number and diversity of predictive models continue to grow, ensuring fair comparisons and reproducible experimental evaluation has become increasingly important. A major source of experimental variability lies in the preprocessing stage, where methodological choices differ substantially across studies. These differences include the selection of external features~\cite{deep,rigamonti2024improving}, strategies for handling missing CGM data~\cite{bg-bert,rodriguez2023prediction}, and the application of filtering, smoothing, and data augmentation techniques~\cite{mhaskar2017deep,de2021adversarial}. Moreover, public datasets often use inconsistent naming conventions for similar variables and include heterogeneous data modalities beyond glucose measurements, requiring dataset-specific preprocessing procedures~\cite{ohio,diatrend}.
% Moreover, due to its sensitive nature, BGC data is often subject to restrictive policies that prevent the sharing of preprocessed data, thus impacting the adoption of common policies for preprocessing and reproducibility.
% Additionally, the sensitive nature of BGC data often subjects it to strict privacy regulations, limiting the sharing of preprocessed datasets. This lack of standardization further complicates the adoption of unified preprocessing protocols and undermines reproducibility in the field.
% These challenges are further exacerbated by the sensitive nature of BGC data, as privacy regulations often prevent the redistribution of preprocessed datasets. Consequently, reproducing published preprocessing pipelines and establishing standardized evaluation protocols become considerably more difficult.
These challenges are further exacerbated by the sensitive nature of BGC data, as privacy regulations often prevent the redistribution of preprocessed datasets. Consequently, reproducing published experiments and establishing standardized evaluation protocols become considerably more difficult, underscoring the need for standardized preprocessing workflows and transparent experimental pipelines.

% Among existing efforts aimed at supporting BGC prediction and benchmarking, \textit{GluPredKit}~\cite{wolff2024glupredkit} provides an open-source framework that integrates preprocessing, model training, and evaluation. Although it represents an important step towards structured model comparison, its flexibility is constrained by a fixed preprocessing workflow and tight integration with specific ML models, limiting the researchers’ ability to adapt the framework to their own datasets and study objectives.
Several software tools have been proposed to support BGC prediction and benchmarking.
Among these, \textit{GluPredKit}~\cite{wolff2024glupredkit} provides an open-source framework integrating preprocessing, model training, and evaluation. Although it represents an important step toward structured benchmarking, its flexibility is constrained by a predefined preprocessing workflow and tight coupling with specific ML models, reducing its adaptability across different datasets and research objectives.

% Other tools instead primarily focus on CGM-derived metric computation and exploratory analysis rather than on configurable preprocessing for predictive modeling. For instance, \textit{iglu}~\cite{broll2021interpreting} offers a comprehensive set of CGM-derived metrics and graphical visualizations to assess glucose control and variability. Similarly, \textit{Glucostats}~\cite{peiro2025glucostats} enables efficient computation and visualization of a wide range of standardized glucose metrics. While both tools are valuable for clinical interpretation and exploratory analysis, they do not provide structured and extensible preprocessing pipelines suitable for end-to-end machine learning workflows.
% Continuous Glucose Monitoring Time Series Data Analysis (\textit{CGMTSA})~\cite{shao2023continuous} explicitly approaches CGM as time-series data, offering dedicated methods for missing data imputation, outlier detection, computation of recommended metrics, and interactive temporal visualizations. Although it strengthens the statistical analysis of CGM time series, it was not conceived as a modular and extensible preprocessing framework for ML pipelines.
Other tools primarily focus on CGM-derived metric computation and exploratory analysis rather than configurable preprocessing for predictive modeling. For instance, \textit{iglu}~\cite{broll2021interpreting} offers a comprehensive set of CGM-derived metrics and visualizations for assessing glucose control and variability, whereas \textit{Glucostats}~\cite{peiro2025glucostats} enables efficient computation and visualization of a wide range of standardized glucose metrics. Similarly, Continuous Glucose Monitoring Time Series Data Analysis (\textit{CGMTSA})~\cite{shao2023continuous} approaches CGM as time-series data by providing methods for missing-data imputation, outlier detection, metric computation, and interactive temporal visualizations. Although these tools provide valuable support for statistical analysis and clinical interpretation, they are not designed as modular and extensible frameworks for reproducible end-to-end machine learning experimentation.

% To overcome these limitations, we present \Tool{}\footnote{\label{github}\faIcon{github}~\url{https://unimib-islab.github.io/glucotune}}, a modular, model-agnostic, and highly configurable framework for end-to-end BGC time series analysis. \Tool{} integrates advanced preprocessing functionalities and a comprehensive library of statistical, ML, and DL models within a unified environment, facilitating both data preparation and the construction of strong baselines for comparison with state-of-the-art methods.
To address these limitations, we present \Tool{}\footnote{\label{github}\faIcon{github}~\url{https://unimib-islab.github.io/glucotune}}, a modular and extensible framework that standardizes the complete workflow for reproducible experimentation with BGC time-series data, from preprocessing to model evaluation. Configurable preprocessing pipelines are fully specified through portable YAML configuration files, enabling experiments to be reproduced directly from the original datasets without requiring the redistribution of sensitive preprocessed data. 

% On the data preparation side, it provides flexible and customizable support for missing data handling, feature selection, glycemic thresholding, data splitting, augmentation, and normalization. These components are designed to streamline the preprocessing phase and promote reproducibility across experiments. The framework is compatible with widely used public datasets, ensuring broad adaptability across different study settings.
% In addition, \Tool{} includes a wide range of ready-to-use statistical, ML, and DL models. Some are specifically designed for diabetes-specific predictive tasks, while others are general-purpose approaches for time series analysis. This integrated model library enables users to experiment with different methodologies within the same environment and to quickly establish strong baselines for rigorous comparison with state-of-the-art methods.
The preprocessing layer supports configurable preprocessing operations, including missing-data handling, feature selection, glycemic thresholding, data partitioning, augmentation, and normalization. \Tool{} includes publicly available datasets through standardized wrappers, providing a consistent interface across heterogeneous data sources. Building upon this preprocessing layer, it provides a unified environment for implementing, training, and evaluating blood glucose prediction models. The framework integrates a curated collection of statistical, ML, and DL models, including both diabetes-specific predictors and general-purpose time-series forecasting methods, while remaining readily extensible to additional datasets, preprocessing strategies, and forecasting models. This architecture enables researchers to establish strong experimental baselines and systematically compare forecasting approaches under consistent preprocessing and evaluation settings.
%
% The framework is designed to be fully extensible, allowing users to seamlessly integrate new datasets and additional models into the existing pipeline. This modular architecture ensures adaptability to diverse research needs while preserving a unified and reproducible experimental workflow.
% To enhance usability, \Tool{} features an intuitive Graphical User Interface (GUI) that supports the entire workflow, from preprocessing configuration to model selection, training, evaluation, and comparison. The interface also includes built-in visualizations to help users explore BGC statistics and inspect results.
% \Tool{} is outlined for researchers and healthcare data scientists working with BGC time series who seek a flexible, extensible, and user-friendly solution to manage the complete analytical pipeline while ensuring transparent and comparable experimental setups. Its modular structure and intuitive GUI make it accessible even to non-developers.
% To demonstrate the versatility of \Tool{}, we present a leaderboard-based evaluation covering both data preprocessing and baseline model construction, enabling systematic and reproducible comparison of state-of-the-art methods within a unified experimental setting. We additionally report the results of a usability study assessing the framework’s accessibility and user experience. These results illustrate \Tool{}’s potential to standardize preprocessing pipelines and improve the reproducibility and comparability of ML research in diabetes care.
%
Its modular architecture facilitates the seamless integration of additional datasets, preprocessing strategies, and forecasting models while preserving a unified and reproducible experimental workflow.

To further improve accessibility, an intuitive Graphical User Interface (GUI) supports the complete workflow, from preprocessing configuration to model selection, training, evaluation, and result comparison, while integrated visualizations facilitate data exploration and interpretation.

% To demonstrate the capabilities of \Tool{}, we present a leaderboard-based evaluation spanning multiple preprocessing configurations and baseline forecasting models, enabling systematic and reproducible comparison under a unified experimental setting. We further assess the usability of \Tool{} through a dedicated user study. Together, these evaluations demonstrate the effectiveness of the framework and its potential to improve the reproducibility and comparability of blood glucose prediction research.
To evaluate the proposed framework, we present a benchmarking leaderboard reporting results across different preprocessing configurations and baseline forecasting models, enabling systematic and reproducible comparison under a unified experimental setting. We further assess the usability of \Tool{} through a dedicated user study. Together, these evaluations demonstrate the effectiveness of the framework and its potential to improve the reproducibility and comparability of blood glucose prediction research.

\section{Methods}\label{methods}
% This section describes the components of \Tool{}, including the configurable preprocessing pipeline, the integrated baseline model construction framework, and the GUI.

% \Tool{} is offered as a set of cross-platform Python packages: data preprocessing, BGC forecasting baselines, and GUI. Its modular design offers flexibility, allowing install and use only the components of interest. It also makes possible the seamless integration (via a standardized interface) of new datasets, preprocessing and forecasting methods, to tailor data preparation and to BGC prediction to specific needs.

% To support reproducibility of experiments the configuration of the preprocessing pipeline is defined via YAML text files. This allows sharing the preprocessing setup without sharing the preprocessed data that is usually forbidden by the terms of use of the datasets due to the sensitive nature of the data. Thanks to this it is possible for other researcher to obtain independently the access to the datasets and replicate the experimental setup.
This section presents the architecture of \Tool{}, including its configurable preprocessing pipeline, integrated framework for baseline modeling and evaluation, and Graphical User Interface (GUI).

\Tool{} is distributed as a collection of cross-platform Python packages covering data preprocessing, baseline forecasting methods, and the GUI. Its modular architecture allows users to install and use only the components relevant to their needs while providing a standardized interface for integrating additional datasets, preprocessing strategies, and forecasting models. This design facilitates straightforward extension of the framework while preserving a unified and reproducible experimental workflow.

To promote reproducible experimentation, preprocessing workflows are fully specified through portable YAML configuration files. This approach allows researchers to share complete preprocessing configurations without redistributing preprocessed datasets, whose redistribution is often restricted by privacy regulations and dataset usage policies. Consequently, other researchers can independently gain access to the original datasets and reproduce the entire experimental setup under identical conditions.

\subsection{Preprocessing Pipeline}
\label{section_2}
\begin{figure}%[tb]
\centering
  \includegraphics[width=\linewidth]{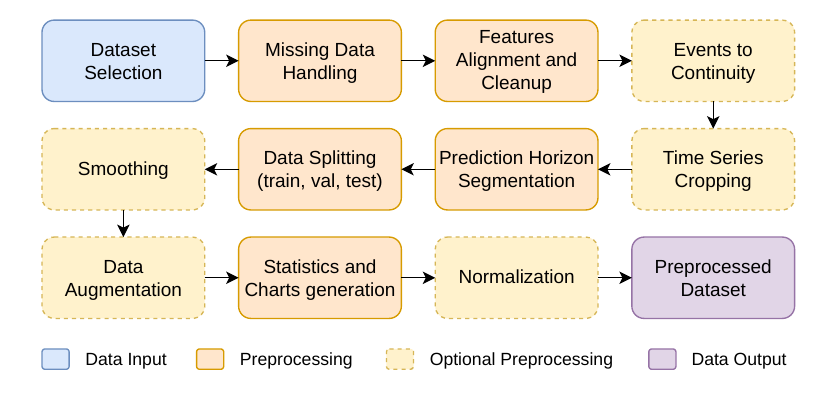}
  % \captionsetup{font=small} 
  \caption{Workflow of \Tool{} data preprocessing; dashed lines indicate optional components that execute only in specific user configurations.}
  % \caption{Workflow of \Tool{}; Dashed lines indicate optional components that execute only under specific conditions (e.g., Bolus-Carbs continuity, available exclusively on the OhioT1DM and DiaTrend datasets when Bolus and/or Carbs are selected as features).}
  \label{fig:GlucoTune_Pipe}
\end{figure}

% As illustrated in Fig. \ref{fig:GlucoTune_Pipe}, this section focuses on the data preparation workflow. The preprocessing pipeline is designed to standardize and streamline BGC time series data, currently supporting four distinct datasets and incorporating a range of preprocessing strategies, integrating key steps from state-of-the-art methods to ensure adaptability and ease of use. 
% The following provides a detailed overview of its workflow.
As illustrated in Fig. \ref{fig:GlucoTune_Pipe}, this section presents the data preparation workflow implemented in \Tool{}. The preprocessing pipeline standardizes blood glucose time-series data through a configurable sequence of preprocessing operations. It currently supports four public datasets and integrates a broad range of preprocessing strategies derived from state-of-the-art methods, providing a flexible and extensible framework for preparing data for blood glucose prediction.
% The following subsections describe each stage of the preprocessing workflow in detail.
The following subsections provide a detailed overview of its workflow.
% \begin{figure}[ht]
% \centering
%   \includegraphics[width=7cm]{images/demo.png}
%   \captionsetup{font=small} 
%   \caption{Workflow of \Tool{}.}
%   \label{fig:GlucoTune_Pipe}
% \end{figure}

\subsubsection{Dataset Selection} 
Currently, the framework supports four major state-of-the-art Type 1 Diabetes (T1D) datasets. (1) OhioT1DM~\cite{ohio} contains eight weeks of data from 12 individuals with T1D released for the 2018 and 2020 challenges. It includes CGM measurements sampled every 5 minutes, insulin delivery records (basal and bolus), self-reported carbohydrate intake, event annotations (e.g., exercise and stress), and optional wearable sensor data such as heart rate and activity signals. 
(2) DiaTrend~\cite{diatrend} features longitudinal data from 54 individuals with T1D. It comprises 27,561 days of CGM data at 5-minute intervals, 8,220 days of insulin pump data (including basal insulin for 17 subjects), bolus doses, carbohydrate intake, pump settings, and detailed demographic and clinical profiles. 
(3) T1DiabetesGranada~\cite{rodriguez2023t1diabetesgranada} is one of the largest and longest open-access longitudinal datasets for type 1 diabetes research. It includes 257,780 days of 15-minute CGM recordings collected over four years from 736 patients
(4) T1DEXI~\cite{riddell2023examining} contains data of 497 adults with type 1 diabetes participating in a four-week structured at-home exercise study. The dataset includes continuous glucose monitoring data sampled every 5 minutes, together with heart rate, insulin dosing, and self-reported exercise and dietary intake collected via a custom smartphone application.

This step also uniforms the naming conventions of available data across datasets to standardize the subsequent processing.

\subsubsection{Missing Data Handling}
During this phase, users can decide whether to handle missing data by ensuring a uniform interval between observations and filling in gaps caused by missing BGC readings via interpolation. 
% This is currently done by linear interpolation.
%Presently, linear interpolation is the only implemented method for imputing missing values.
Additionally, users can specify the maximum allowable gap in CGM data before splitting the time series. If a missing interval exceeds this threshold, the time series will be segmented. % accordingly.

\subsubsection{Feature Alignment and Cleanup} Depending on the chosen dataset, users can pick different features. 
First, \Tool{} synchronizes the selected features with CGM readings, ensuring that each BGC timestamp corresponds to observations within a maximum delay defined by the observation frequency of the dataset.
This process eliminates potential information gaps before the BGC measurements, improving data consistency. 
Next, it applies filtering techniques to both features and CGM readings to correct discrepancies and remove outliers caused by recording errors. 
%Finally, given that some features may span a more limited period than CGM readings, the system crops all-time series to ensure that all features remain temporally consistent. % with the CGM data.

\subsubsection{Events to Continuity}
%\DELETED{For instance, T1DiabetesGranada provides only DateTime and CGM measurements. In contrast, multimodal datasets DiaTrend and OhioT1DM, support additional features such as bolus insulin doses and carbohydrate intake.}
% The toolkit alerts the user if a selected feature is not available for certain subjects.
If bolus insulin or carbohydrate intake data is selected, users can enable an optional continuity process to ensure a more consistent representation of these features, following the approach in~\cite{butt2023feature}. %Specifically, bolus insulin data is processed using an exponential decay curve to estimate insulin on board (IOB), while carbohydrate intake is modeled with a custom function that simulates absorption rates over time, accounting for post-ingestion fluctuations.

\subsubsection{Time Series Cropping}
Given that some features may span a more limited period than CGM readings, if the users selected features that are present in a limited time span the system crops all-time series to ensure that all features remain temporally consistent. % with the CGM data.

\subsubsection{Prediction Horizon Segmentation}
The time series is divided into samples according to the user-defined PH, the selected number of past observations (input), and the specified number of future observations (target).
For example, if the PH is 30 minutes and CGM readings occur every 5 mins, each time series segment consists of 30 samples (24 input + 6 target).

% Next, users can choose how to split the time series and define the percentage of data allocated to the train, validation, and test sets. The toolkit supports three strategies: (1) Temporal split, where data is divided chronologically to train on past data, validate on intermediate data, and test on future data for a realistic performance assessment; (2) Stratified split, which ensures a balanced distribution of hypo-/normo-/hyper-glycemic events across train, validation, and test sets; and (3) Hybrid split, combining both approaches. % by selecting the test set temporally while stratifying the training and validation sets to maintain event balance.
% This structured approach provides flexibility in dataset preparation while maintaining both temporal consistency and event distribution.

\subsubsection{Data Splitting}
% This step generates the train, validation, and tests data splits of the dataset.
Users can choose how to split the time series and define the percentage of data allocated to the train, validation, and test sets. The framework supports three strategies: (1) \textit{Temporal split}, where data is divided chronologically to train on past data, validate on intermediate data, and test on future data for a realistic performance assessment; (2) \textit{Stratified split}, which ensures a balanced distribution of hypo-/normo-/hyper-glycemic events across train, validation, and test sets; and (3) \textit{Hybrid split}, combining both approaches. 
For this stage, \Tool{} allows users to customize hypoglycemia and hyperglycemia thresholds which impacts the data distribution in the splits. 
This flexibility supports varying definitions in research. 
For example, while some studies classify hypoglycemia as below 70 mg/dL, others use 80 mg/dL. Similarly, hyperglycemia can be defined as above 180 mg/dL or 250 mg/dL, depending on study criteria (e.g., \cite{deng2021deep,bg-bert}).

\subsubsection{Smoothing}
Users can optionally apply smoothing filters to CGM data to reduce noise. % and enhance data quality. 
The framework uses a Gaussian filter with configurable sigma value, allowing fine control over the smoothing process. %However, its modular design enables the seamless integration of additional smoothing techniques as needed.

\subsubsection{Data Augmentation}
\Tool{} incorporates data augmentation techniques which can be optionally used to address class imbalance. %, particularly in adverse glucose events. 
% Currently, it supports SMOTE (Synthetic Minority Over-sampling Technique~\cite{chawla2002smote}), which users can customize to align with their study objectives. They can apply a multiplier factor or specify the target number of hypoglycemia and hyperglycemia events, as well as choose whether to apply SMOTE globally across the dataset or individually for each subject.
It currently supports Synthetic Minority Over-sampling Technique (SMOTE)~\cite{chawla2002smote}, which users can customize by applying a multiplier factor or specifying the target number of hypo- and hyperglycemia events. It can be applied globally across the dataset or individually for each subject.

\subsubsection{Statistics and Charts Generation}
At this stage the preprocessing is mostly complete and statistics on the data are extracted as well as plots about the data split, glucose data, and sample time series are generated.

\subsubsection{Normalization}
Lastly, users can choose whether to apply data normalization, with the option to select Min-Max Normalization or Max Normalization. % according to their requirements.

\subsection{Model Library and Baseline Construction}
Once the preprocessing stage is completed, \Tool{} enables users to construct baseline models, facilitating straightforward comparison with state-of-the-art approaches. In this section, we describe the model library and the baseline construction module of the framework. This functionality streamlines experimental evaluation, allowing custom methods to be benchmarked within a unified and reproducible environment.

\begin{figure}[tb]
\centering
  \includegraphics[width=0.9\linewidth]{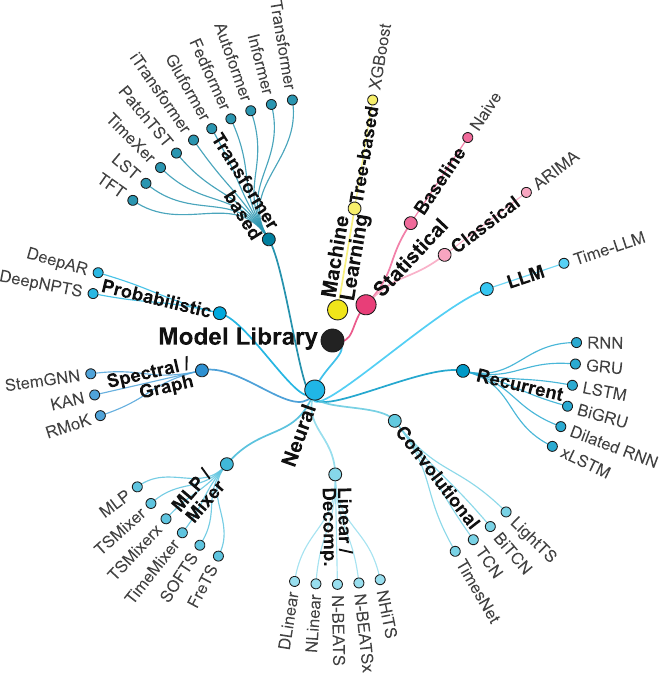}
    \caption{General overview of the models currently available in the \Tool{} model library.}
  \label{fig:GlucoTune_Models}
\end{figure}

The current model library spans multiple methodological families, including classical statistical approaches, traditional machine learning algorithms, and a wide range of neural architectures. Specifically, it comprises classical statistical models (e.g., Naive, ARIMA), tree-based machine learning methods (e.g., XGBoost), and several neural model families such as recurrent networks (e.g., RNN, LSTM, GRU, BiGRU), convolutional and temporal convolutional networks (e.g., TCN), Transformer-based architectures (e.g., Informer, Autoformer, FEDformer), linear and decomposition-based models (e.g., N-BEATS), as well as more recent approaches including MLP-Mixer variants, probabilistic models, graph-based models, and LLM-based approaches.
Most of the implemented models have been selected for their relevance to blood glucose prediction tasks. In addition, the library includes well-established general-purpose time series forecasting models adapted to the glucose prediction setting, thereby broadening the range of comparative baselines.
Fig.~\ref{fig:GlucoTune_Models} illustrates the hierarchical organization of the model library, highlighting the different methodological families and their associated implementations. 

As with the preprocessing component, is designed to be highly modular, extensible, and flexible. Its architecture enables the seamless integration of new models with minimal effort, allowing the framework to evolve alongside advances in time series forecasting and glucose prediction methodologies.

% \TODO{Da aggiungere qualche dettaglio in più?}
% % \TODO{Da aggiungere i FLOPS, in tabella e prepara una figura stile bubble plot FLOPS, F1 (sen hypo/hyper), num_params} 
% \TODO{Da aumentare un po' di più la dimensione delle bubble in base al numero di parametri}
% \TODO{Da aumentare font figura 2}

\subsection{Graphical User Interface}
%For ease of use, we provide a graphical user interface that allows users to configure, run, and export the preprocessing setup . 
The GUI is available as a separate cross-platform Python package, making it an optional companion to \Tool{} (see Fig.~\ref{fig:GlucoTune_GUI}). While \Tool{} is fully operable via the command-line interface, the GUI enhances accessibility for non-developers by guiding users through the whole workflow. 
Users can start with the preprocessing pipeline, selecting and customizing the configuration that best fits their experimental needs. They may then export the selected configuration for future reuse or sharing, or directly execute the newly defined strategy. The GUI also supports importing, modifying, and running existing configuration files.
After completing the preprocessing step, users can continue within the \Tool{} GUI to build a custom baseline. The framework supports the selection and configuration of different modeling approaches within the same environment, facilitating straightforward benchmarking against state-of-the-art methods under consistent preprocessing and evaluation settings.

\subsubsection{Data visualization functionality in the GUI}
% To help users better understand their data before and after preprocessing, the GUI also includes a page with basic visualizations. These plots provide a clear overview of key metrics, such as glucose time series trends, the distribution of readings across predefined glucose ranges, and overall data availability over the observation period.  (see Figure~\ref{fig:GlucoTune_Viz}). These visual insights offer an intuitive starting point for exploring and validating glucose data, enhancing the transparency and effectiveness of the preprocessing workflow.

\begin{figure}[tb]
\centering

    \begin{subfigure}[b]{0.3\textwidth}
         \centering
         \includegraphics[width=\textwidth]{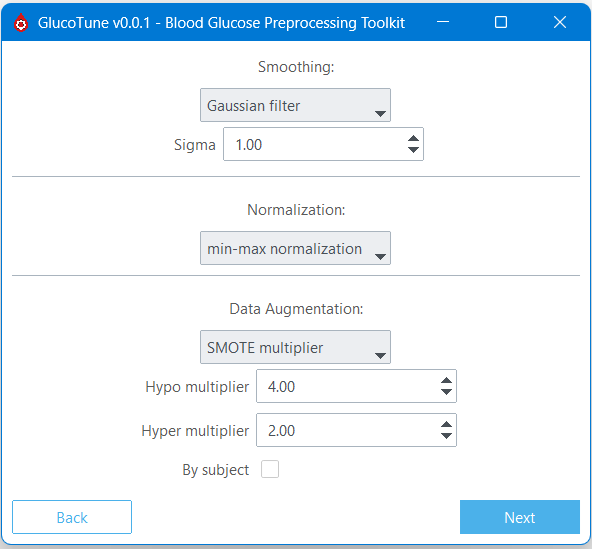}
         \caption{}
    \end{subfigure}\\
     % \hfill
    \vspace{0.5em}
    \begin{subfigure}[b]{0.3\textwidth}
         \centering
         \includegraphics[width=\textwidth]{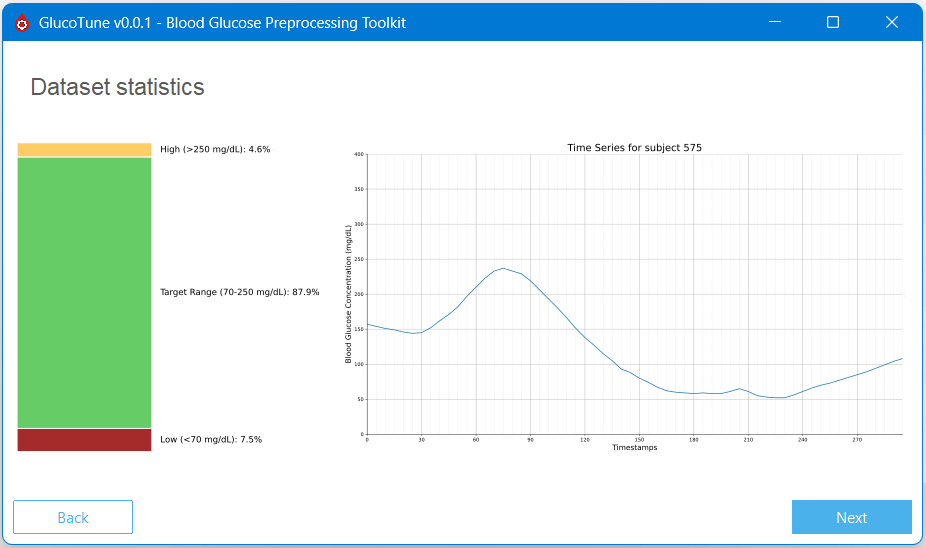}
         \caption{}
     \end{subfigure}

\caption{Graphical User Interface of \Tool{}.}
\label{fig:GlucoTune_GUI}
\end{figure}

To support data understanding before and after preprocessing, the GUI includes a visualization page that displays glucose trends, distribution across predefined glucose ranges, and data availability over time (Fig.~\ref{fig:GlucoTune_GUI}). These plots offer an intuitive overview that aids in interpreting and validating glucose data.

% \begin{figure}[b]
% \centering
%   \includegraphics[width=0.8\linewidth]{images/GlucoTuneGUI_1.png}
%   % \captionsetup{font=small} 
%   \caption{Graphical User Interface of \Tool{}.}
%   \label{fig:GlucoTune_GUI}
% \end{figure}
% \begin{figure}[b]
% \centering
%   \includegraphics[width=0.8\linewidth]{images/GlucoTune1.png}
%   % \captionsetup{font=small} 
%   \caption{Graphical User Interface of \Tool{}.}
%   \label{fig:GlucoTune_GUI}
% \end{figure}

\section{Results}\label{results}
% \subsection{Assessment of \Tool{}} 
We assess \Tool{} through a predictive modeling benchmark and a usability study, highlighting its technical utility and user experience.

\subsection{Predictive Modeling Benchmark}
To validate the effectiveness of \Tool{}, including both preprocessing pipeline and integrated model library, we design a unified benchmarking framework covering all currently supported models.
The evaluation begins by preprocessing the selected datasets using the configurable pipeline of \Tool{}, ensuring consistency across experiments.

We conduct the analysis on two representative datasets among those currently supported, namely OhioT1DM and DiaTrend, considering PH of 30 and 60 minutes, and comparing configurations with and without SMOTE-based data augmentation.
All models are trained and evaluated under the same protocol to ensure fair comparison across heterogeneous families: we adopt a temporal split, apply max normalization, and optimize using Mean Squared Error (MSE) loss. This setup reduces variability due to preprocessing and training choices, helping isolate differences attributable to the models.

The full leaderboard\footnote{\url{https://unimib-islab.github.io/glucotune/leaderboard}} reports a comprehensive set of evaluation metrics commonly adopted in blood glucose prediction, including Root Mean Squared Error (RMSE, overall prediction error), Mean Absolute Error (MAE, average absolute deviation), Mean Absolute Relative Difference (MARD, relative prediction error), Time Gain (TG, temporal prediction advantage), sensitivity and specificity to hypo- and hyperglycemic events (Sen$_{hypo}$, Sen$_{hyper}$, Spec$_{hypo}$, Spec$_{hyper}$), Pearson Correlation Coefficient (PCC, linear agreement), Prediction in clinically critical zones D and E (PDE, probability of clinically wrong treatment)~\cite{neumann2025data}, and time lag ($\tau_{lag}$, systematic prediction delay)~\cite{zhu2020dilated}.
These metrics are jointly considered to provide a standardized and comprehensive evaluation, addressing the fragmented use of individual metrics in prior work, and capturing complementary aspects of predictive performance, including accuracy, temporal alignment, and clinical relevance.

\begin{table*}[t!]
\centering
\caption{Benchmarking of forecasting performance and computational complexity across datasets, forecasting models, and prediction horizons.}
\label{tab:forecasting_results}
\resizebox{\linewidth}{!}{
\begin{tabular}{c l l c c c cccccccc cccccccc}
\toprule
& & & & & \multicolumn{8}{c}{PH = 30 min} & \multicolumn{8}{c}{PH = 60 min} \\
\cmidrule(lr){7-14} \cmidrule(lr){15-22}

Dataset & Model & Type & Aug. & Params. & FLOPs &
RMSE $\downarrow$ & MAE $\downarrow$ & MARD $\downarrow$ & TG $\uparrow$ & Sen$_{hypo}$ $\uparrow$ & Sen$_{hyper}$ $\uparrow$ & Spec$_{hypo}$ $\uparrow$ & Spec$_{hyper}$ $\uparrow$ &
RMSE $\downarrow$ & MAE $\downarrow$ & MARD $\downarrow$ & TG $\uparrow$ & Sen$_{hypo}$ $\uparrow$ & Sen$_{hyper}$ $\uparrow$ & Spec$_{hypo}$ $\uparrow$ & Spec$_{hyper}$ $\uparrow$\\
& & & & PH=30 min & PH=30 min &
(mg/dL) & (mg/dL) & (\%) & (min) & (\%) & (\%) & (\%) & (\%) &
(mg/dL) & (mg/dL) & (\%) & (min) & (\%) & (\%) & (\%) & (\%) \\

\midrule
\multirow{10}{*}{\rotatebox[origin=c]{90}{OhioT1DM}}
& Naive  & \multirow{2}{*}{Statistical} & -- & \phantom{00}0\phantom{K} & \phantom{00}0\phantom{K} 
& 16.18 & 10.65 & 7.10 & \phantom{0}7.61 & 74.13 & 85.91 & 99.25 & 98.56 & 25.74 & 17.16 & 11.44 & \phantom{0}8.33 & 60.18 & 75.96 & 98.88 & 97.70 \\
& ARIMA & & -- & \phantom{00}3\phantom{K} & \phantom{00}7K
& 16.28 & \phantom{0}9.31 & 6.15 & 14.13 & \textbf{86.17} & \textbf{90.67} & 98.75 & 98.24 & 25.83 & 15.67 & 10.35 & 21.05 & 74.70 & \textbf{82.41} & 98.13 & 97.21 \\
\cmidrule(lr){2-22}
& XGBoost & ML & -- & 115K & \phantom{00}37K
& 13.61 & \phantom{0}8.64 & 5.83 & 15.58 & 57.62 & 84.96 & \textbf{99.67} & 99.09 & 28.76 & 18.75 & 11.12 & 27.85 & 27.34 & 80.90 & \textbf{99.89} & 96.65 \\
\cmidrule(lr){2-22}
& Transformer & \multirow{7}{*}{Neural} & \checkmark & 293K & \phantom{0}14M
& 13.17 & \phantom{0}8.29 & 5.59 & \textbf{17.15} & 72.96 & 87.95 & 99.39 & 98.92 & 22.07 & 14.57 & \phantom{0}9.92 & 28.82 & 34.07 & 73.22 & 99.71 & 98.63 \\
& NBEATS & & \checkmark & \phantom{00}3M & \phantom{00}5M
& 13.28 & \phantom{0}8.28 & 5.56 & \textbf{17.15} & 71.13 & 86.64 & 99.48 & 99.06 & 22.40 & 14.89 & 10.27 & 27.62 & 26.10 & 74.27 & 99.80 & 98.49 \\
& BiGRU & & \checkmark & 702K & \phantom{0}33M
& \textbf{12.96} & \textbf{\phantom{0}8.15} & \textbf{5.46} & 17.10 & 70.41 & 85.87 & 99.52 & \textbf{99.22} & \textbf{21.42} & \textbf{13.86} & \textbf{\phantom{0}9.28} & \textbf{30.26} & 35.42 & 74.74 & 99.71 & \textbf{98.75} \\
& SOFTS & & \checkmark & \phantom{00}7M & \phantom{0}14M
& 13.56 & \phantom{0}8.48 & 5.65 & 15.16 & 81.45 & 88.95 & 99.17 & 98.88 & 22.73 & 14.74 & \phantom{0}9.84 & 23.83 & \textbf{62.31} & 78.72 & 98.88 & 98.17 \\
& TCN & & \checkmark & 150K & \phantom{00}9M
& 13.45 & \phantom{0}8.50 & 5.71 & 16.33 & 70.08 & 86.40 & 99.48 & 99.05 & 21.99 & 14.70 & \phantom{0}9.91 & 28.27 & 33.97 & 72.25 & 99.70 & 98.57 \\
& KAN & & \checkmark & 369K & 645K
& 13.61 & \phantom{0}8.48 & 5.74 & 15.59 & 66.42 & 86.29 & 99.51 & 99.07 & 22.20 & 14.50 & \phantom{0}9.78 & 27.90 & 45.04 & 72.78 & 99.49 & 98.67 \\
& TimeLLM & & -- & 177M & \phantom{0}99G
& 14.09 & \phantom{0}9.29 & 6.21 & 12.45 & 74.47 & 86.45 & 99.39 & 98.93 & 22.88 & 15.33 & 10.27 & 22.69 & 51.43 & 73.65 & 99.25 & 98.50 \\
\midrule
\multirow{10}{*}{\rotatebox[origin=c]{90}{DiaTrend}}
& Naive & \multirow{2}{*}{Statistical} & -- & \phantom{00}0\phantom{K} & \phantom{00}0\phantom{K}
& 21.30 & 13.87 & 8.05 & \phantom{0}7.89 & 60.76 & 88.18 & 99.40 & 96.87 & 33.91 & 22.30 & 12.94 & \phantom{0}8.99 & 46.13 & 80.83 & 99.18 & 94.93 \\
& ARIMA & & -- & \phantom{00}3\phantom{K} & \phantom{00}7K
& 21.78 & 12.42 & 7.13 & 13.97 & \textbf{79.40} & \textbf{91.81} & 98.80 & 96.62 & 34.14 & 20.78 & 11.91 & 20.24 & \textbf{64.75} & \textbf{85.38} & 98.27 & 94.74 \\
\cmidrule(lr){2-22}
& XGBoost & ML & -- & 148K & \phantom{0}48K
& 17.43 & 11.10 & 6.53 & 15.64 & 47.92 & 89.76 & 99.80 & 97.73 & 28.76 & 18.75 & 11.12 & 27.85 & 27.34 & 80.90 & 99.89 & 96.65 \\
\cmidrule(lr){2-22}
& Transformer & \multirow{7}{*}{Neural} & -- & 293K & \phantom{0}14M
& 16.94 & \textbf{10.69} & \textbf{6.27} & \textbf{16.98} & 58.88 & 90.72 & 99.66 & 97.59 & \textbf{28.21} & \textbf{18.26} & \textbf{10.72} & \textbf{31.24} & 33.93 & 81.52 & 99.79 & 96.63 \\
& NBEATS & & \checkmark & \phantom{00}2M & \phantom{00}5M
& 17.07 & 10.76 & 6.30 & 16.29 & 55.38 & 90.54 & 99.73 & 97.62 & 28.54 & 18.45 & 10.86 & 29.47 & 30.52 & 81.94 & 99.86 & 96.48 \\
& BiGRU & & \checkmark & 702K & \phantom{0}33M
& \textbf{16.92} & 10.74 &  6.33 & 16.70 & 50.08 & 90.11 & \textbf{99.81} & \textbf{97.77} & 28.22 & 18.35 & 10.81 & 30.78 & 25.83 & 80.79 & \textbf{99.92} & \textbf{96.88} \\
& SOFTS & & \checkmark & \phantom{00}7M & \phantom{0}14M
& 17.75 & 11.17 & 6.49 & 14.16 & 73.75 & 91.35 & 99.29 & 97.23 & 29.84 & 19.29 & 11.29 & 24.21 & 52.20 & 83.77 & 99.19 & 95.60 \\
& TCN & & \checkmark & 150K & \phantom{00}9M
& 16.97 & 10.80 & 6.35 & 16.65 & 54.63 & 90.90 & 99.73 & 97.50 & 28.26 & 18.36 & 10.85 & 30.23 & 28.60 & 82.20 & 99.87 & 96.43 \\
& KAN & & \checkmark & 369K & 645K
& 17.13 & 10.86 & 6.38 & 16.58 & 54.37 & 90.52 & 99.74 & 97.60 & 28.50 & 18.54 & 10.90 & 31.15 & 29.41 & 81.23 & 99.85 & 96.60 \\
& TimeLLM & & -- & 177M & \phantom{0}99G
& 18.51 & 12.31 & 7.16 & 12.16 & 59.70 & 88.70 & 99.53 & 97.39 & 30.50 & 20.74 & 12.21 & 23.61 & 34.31 & 78.78 & 99.67 & 96.27 \\
\bottomrule
\end{tabular}
}
\end{table*}

\begin{figure*}[tb]
\centering

    \begin{subfigure}[b]{0.3\textwidth}
         \centering
         \includegraphics[width=\textwidth]{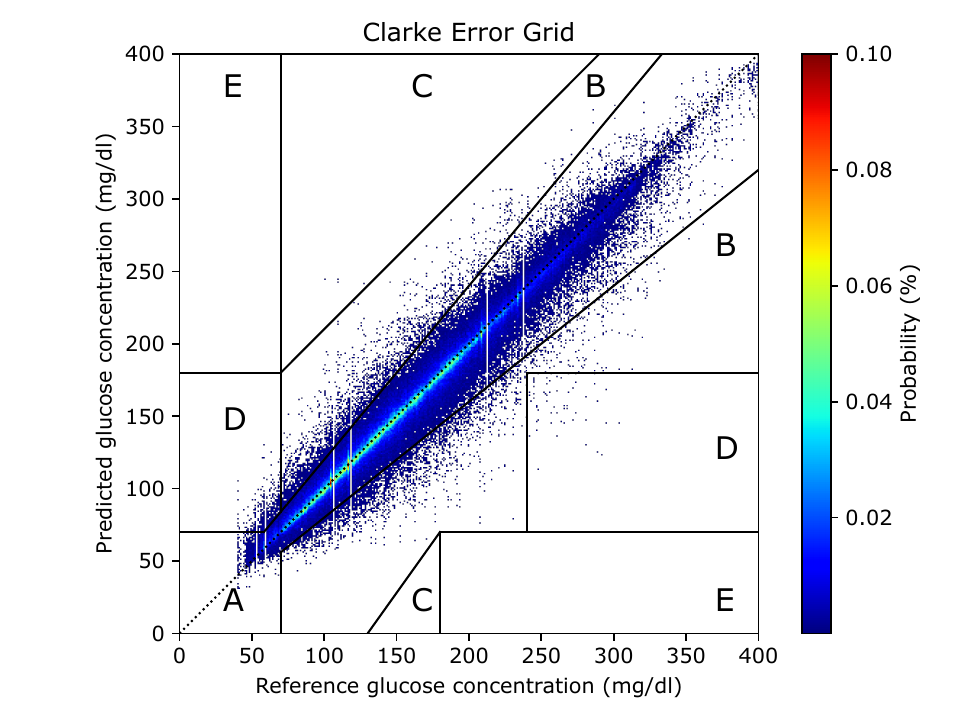}
         \caption{Clarke Error Grid}
     \end{subfigure}
     \hfill
    \begin{subfigure}[b]{0.3\textwidth}
         \centering
         \includegraphics[width=\textwidth]{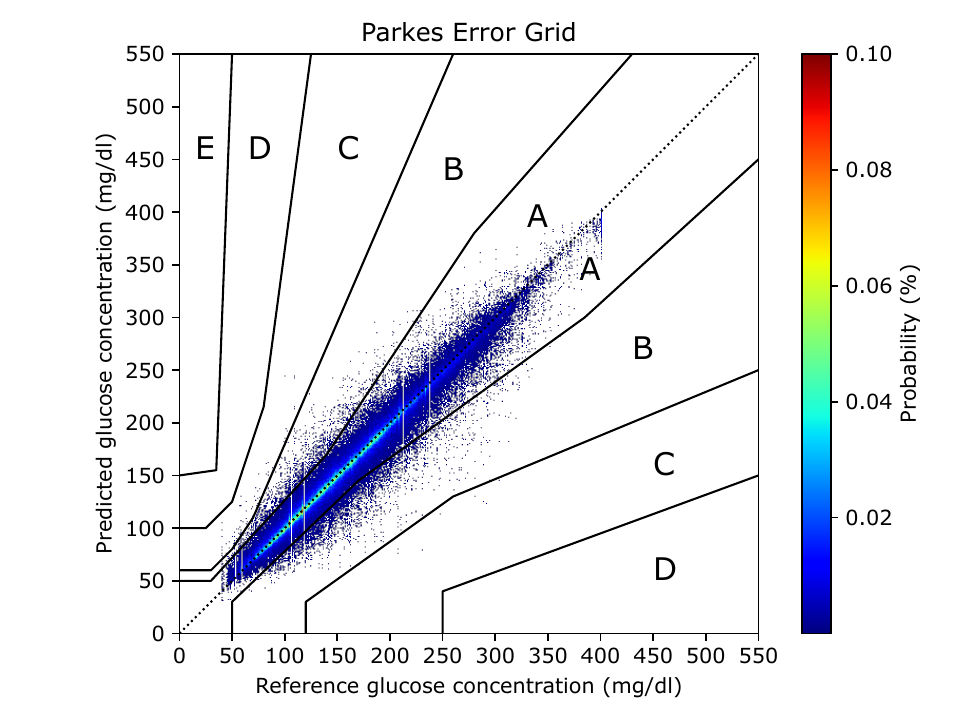}
         \caption{Parkes Error Grid}
     \end{subfigure}
     \hfill
    \begin{subfigure}[b]{0.3\textwidth}
         \centering
         \includegraphics[width=\textwidth]{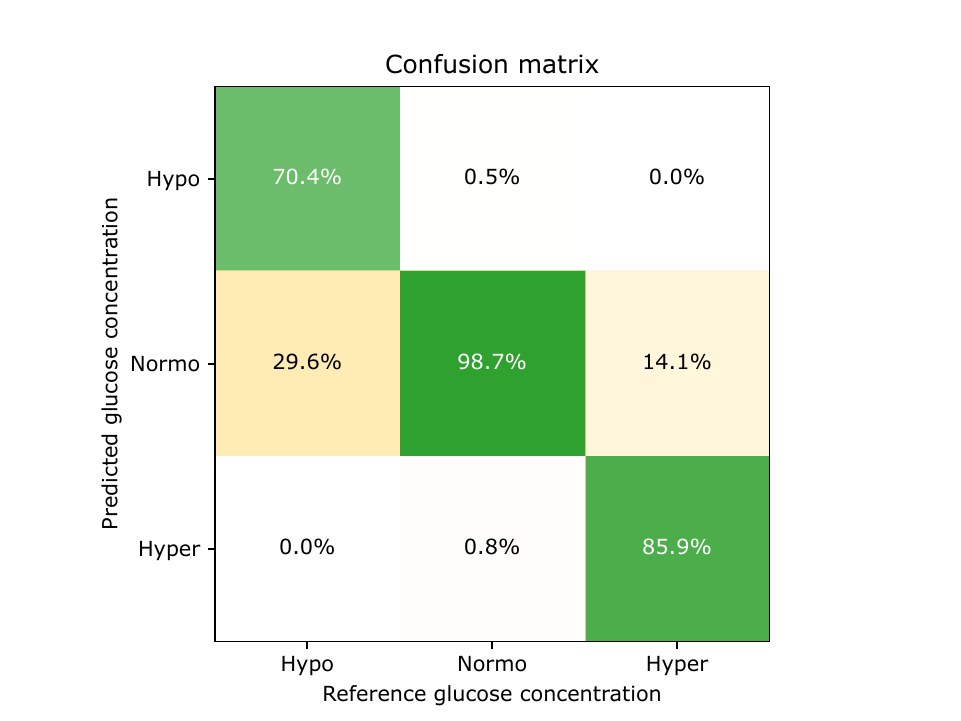}
         \caption{Confusion Matrix}
     \end{subfigure}
     \hfill

\caption{Plots generated by the \Tool{} Baselines module for the BiGRU model on the OhioT1DM dataset with a 30-minute prediction horizon and data augmentation enabled.}
\label{fig:grids-plots}
\end{figure*}

For clarity, Table~\ref{tab:forecasting_results} reports the top 10 models using a subset of widely adopted metrics (RMSE, MAE, MARD, TG, sensitivity and specificity to hypo- and hyperglycemia), providing a concise yet informative summary of predictive performance across datasets and prediction horizons~\cite{bg-bert, barbato2025lightweight}. To facilitate comparison across different classes of forecasting methods, the table also reports the number of trainable parameters and Floating Point Operations (FLOPs), allowing readers to jointly assess predictive accuracy and computational complexity. Model selection is performed at the family level: all statistical and traditional machine learning models are included due to their limited number, while neural architectures are ranked within each family using a composite score derived from normalized key metrics (RMSE, TG, and sensitivity), retaining only the top-performing models.
In this case, model selection is performed at the family level: all statistical and traditional machine learning models are included due to their limited number, while neural architectures are ranked within each family using a composite score derived from normalized key metrics (RMSE, TG, and sensitivity), retaining only the top-performing models.

In addition to quantitative performance metrics, \Tool{} automatically generates a set of visual evaluation plots (Fig.~\ref{fig:grids-plots}) to facilitate qualitative comparison of forecasting methods. Among these, the Clarke Error Grid~\cite{ega} and the Parkes Error Grid~\cite{pfutzner2013technical} assess the clinical relevance of glucose predictions by classifying each forecast into risk zones (A--E). Ideally, predictions should fall within Zone A, indicating clinically accurate estimates, whereas Zones B and C correspond to progressively higher clinical risk and Zones D and E represent potentially dangerous predictions that may lead to inappropriate treatment decisions. In addition, the confusion matrix summarizes the percentages of correctly and incorrectly classified glycemic events, providing further insight into the classification performance of each forecasting method.

\begin{figure*}[htb]
\centering

    \begin{subfigure}[b]{0.495\textwidth}
         \centering
         \includegraphics[width=\textwidth]{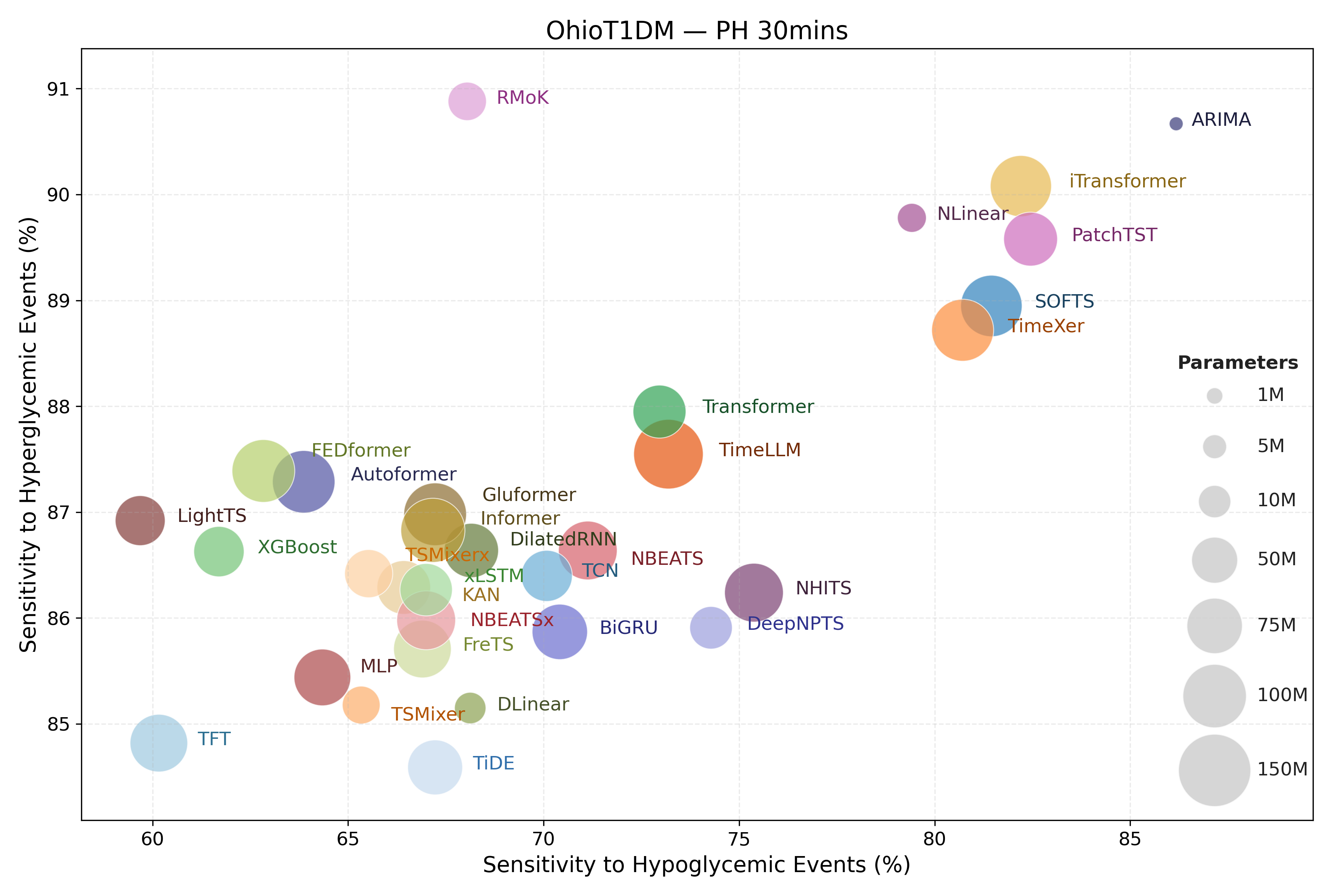}
         \caption{}
     \end{subfigure}
     \hfill
    \begin{subfigure}[b]{0.495\textwidth}
         \centering
         \includegraphics[width=\textwidth]{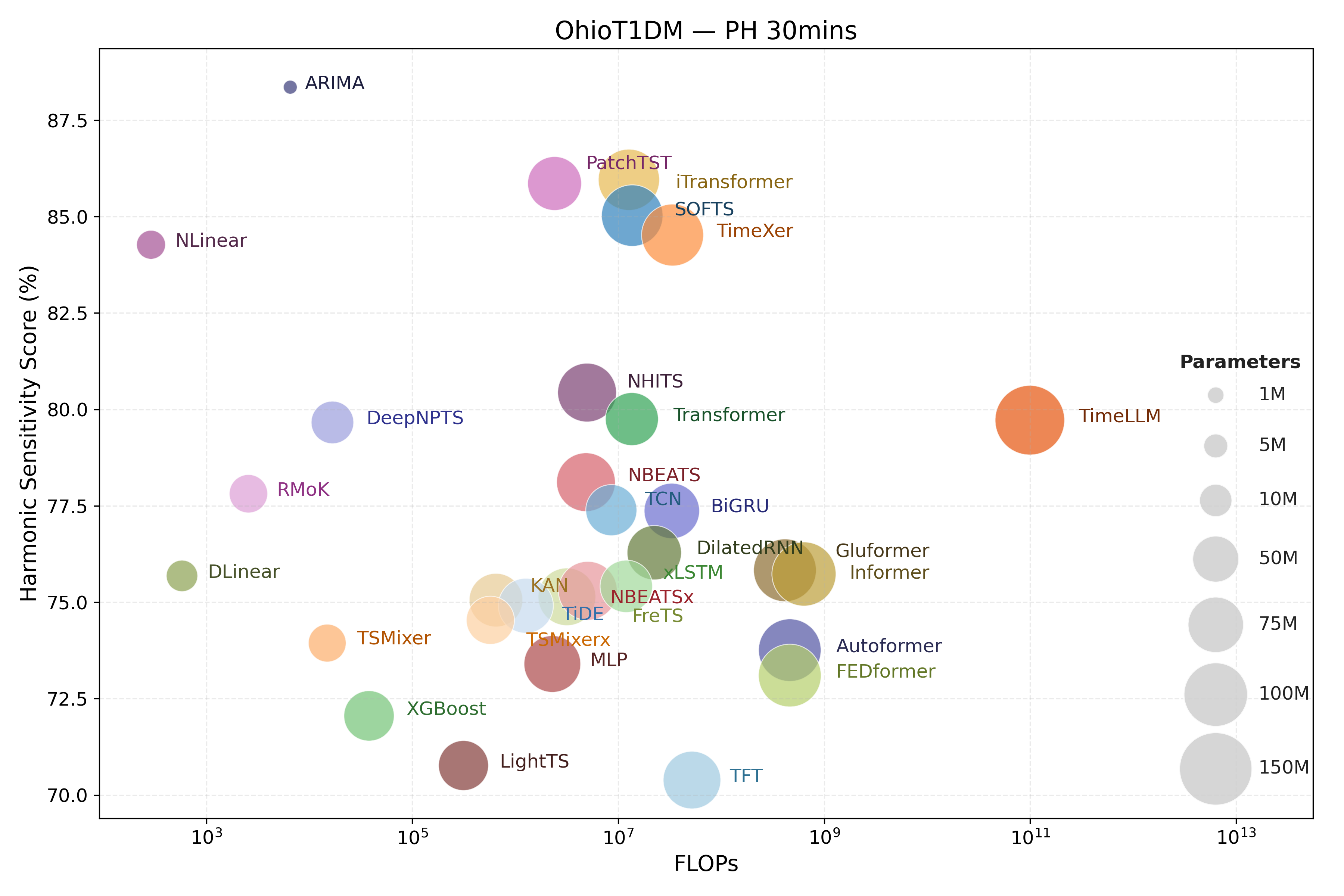}
         \caption{}
     \end{subfigure}

    % \\
    
    \begin{subfigure}[b]{0.495\textwidth}
         \centering
         \includegraphics[width=\textwidth]{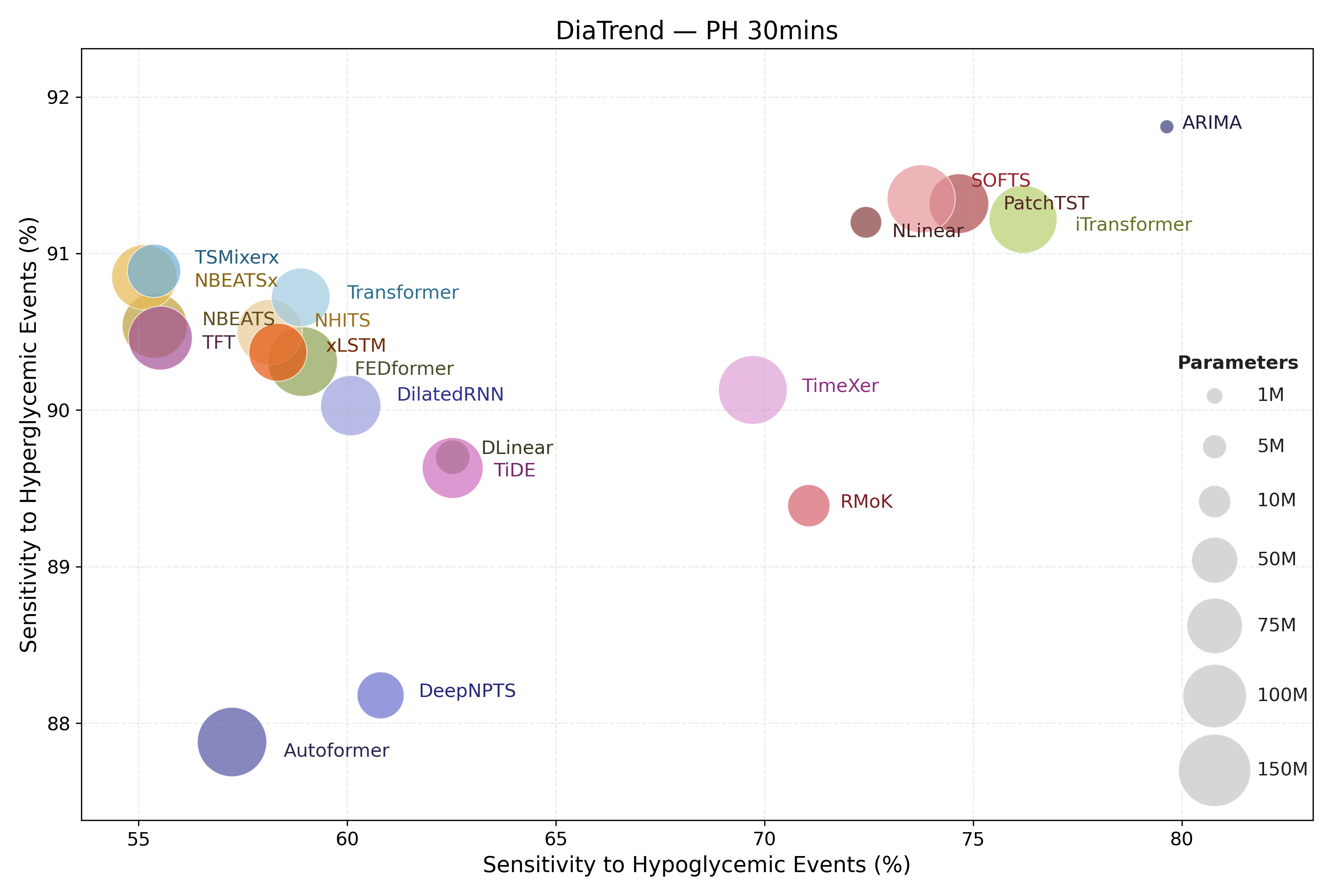}
         \caption{}
     \end{subfigure}
     \hfill
    \begin{subfigure}[b]{0.495\textwidth}
         \centering
         \includegraphics[width=\textwidth]{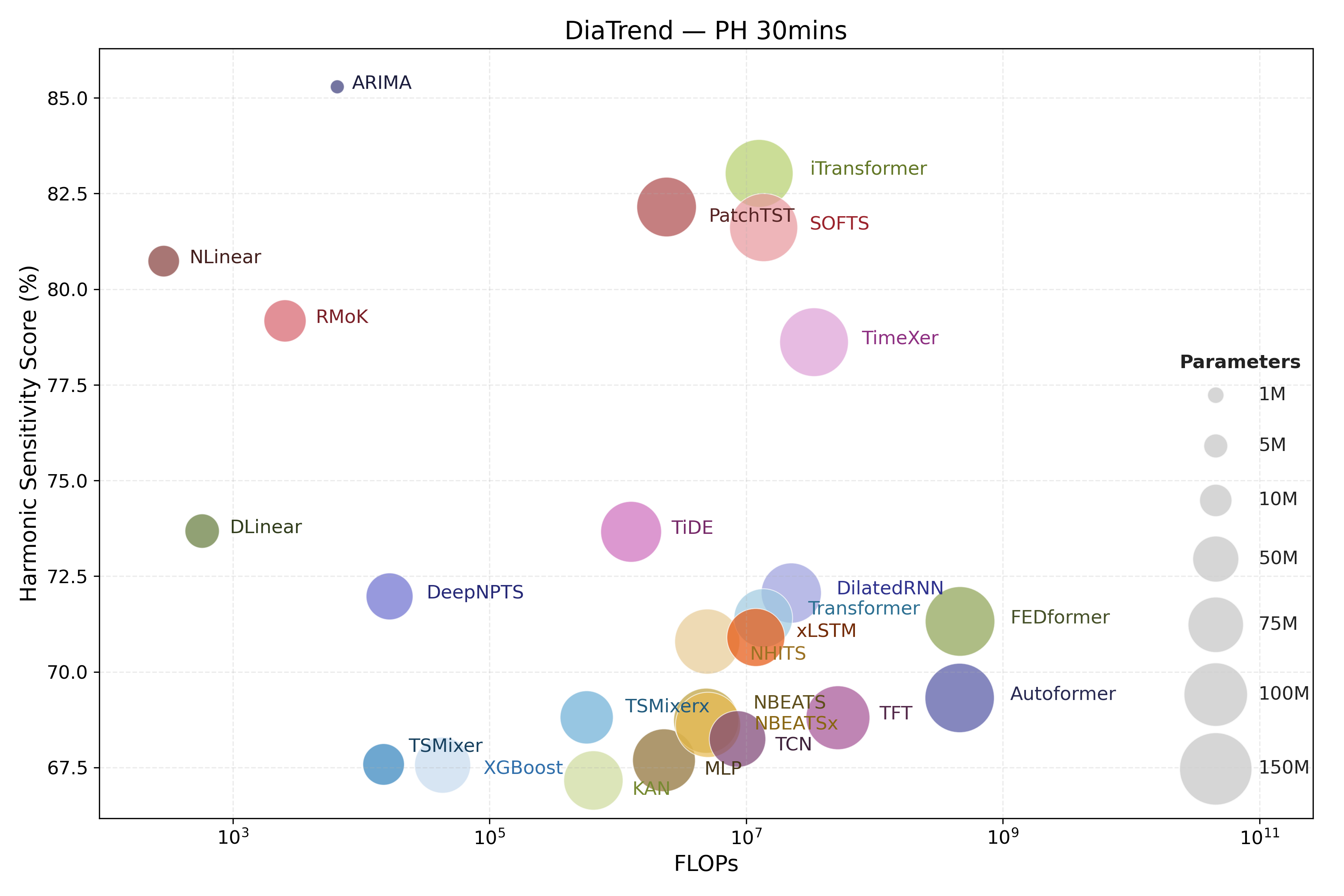}
         \caption{}
     \end{subfigure}

\caption{Bubble plots illustrating the relationship between sensitivity to adverse glycemic events and model complexity, quantified by the number of trainable parameters (a, c) and FLOPs (b, d), on the OhioT1DM and DiaTrend datasets.}
\label{fig:flops-plots}
\end{figure*}

To further analyze the benchmarking results, we employ bubble plots that relate clinically relevant performance metrics to model complexity, quantified by the number of trainable parameters and FLOPs, across all evaluated forecasting models. These visualizations facilitate the analysis of the trade-off between predictive performance and computational cost (Fig.~\ref{fig:flops-plots}).

\subsection{System usability study}
To assess usability of GlucoTune's GUI, we conducted a study with participants from the University of Milano-Bicocca. Each user performed a standardized preprocessing task using GlucoTune, involving the OhioT1DM dataset, selection of CGM, bolus, and carbohydrate features, smoothing ($\sigma$=1), max normalization, and a 30-minute PH. All users performed the task in the same controlled environment without external disturbances, no time limit was imposed. %Remaining parameters were left to user discretion.
After the task, participants completed a questionnaire combining the standard System Usability Scale (SUS, questions 1–10)~\cite{brooke1996sus} with five custom items targeting GlucoTune-specific features:
% \begin{enumerate}
%     \item I found it easy to configure GlucoTune according to the requirements.
%     \item I found it useful to preprocess the dataset without writing any code.
%     \item The preprocessing steps were clearly explained and easy to follow.
%     \item The graphical interface supported me well throughout the task.
%     \item I would feel confident repeating this preprocessing workflow in the future.
% \end{enumerate}
(1) I found it easy to configure GlucoTune according to the requirements; (2) I found it useful to preprocess the dataset without writing any code; (3) The preprocessing steps were clearly explained and easy to follow; (4) The graphical interface supported me well throughout the task; (5) I would feel confident repeating this preprocessing workflow in the future.
A total of 11 participants completed the study. The results indicate that \Tool{} is highly usable, with an average SUS score of 80.91 and an additional 22.18 out of 25 from the custom questions.
The average usability score set from a study on 500 systems is 68~\cite{sauro2011measuring}.
Participants described the tool as intuitive, appreciated the no-code approach, and felt confident completing the task autonomously.
A few suggestions were made to improve usability: users recommended adding tooltips to explain key features, making the interface more self-explanatory and accessible. These insights will inform future refinements, aimed at enhancing user guidance through clearer, more interactive support in \Tool{}.

% SUS SCORE (Sulle dieci domande 80.91 sulle 5 domande 22.18)

\section{Discussion}\label{discussion}
The experimental evaluation demonstrates the practical value of \Tool{} as a unified framework for reproducible blood glucose prediction research. Rather than emphasizing differences between forecasting models, the benchmarking results show that predictive performance is strongly influenced by the experimental setting, including dataset characteristics, preprocessing choices, and prediction horizon. These findings highlight the importance of standardized preprocessing and evaluation protocols to enable fair and meaningful comparison across studies.

By standardizing the complete experimental workflow, from preprocessing to model evaluation, \Tool{} reduces the variability introduced by heterogeneous data preparation and implementation-specific preprocessing and evaluation choices. Consequently, differences in predictive performance can be interpreted with greater confidence as reflecting the forecasting models themselves rather than inconsistencies in the experimental pipeline. This directly addresses one of the main challenges in blood glucose prediction research, where heterogeneous preprocessing procedures and restrictions on sharing preprocessed datasets often hinder reproducibility and objective benchmarking.

Beyond improving reproducibility, \Tool{} simplifies the development and evaluation of forecasting methods through configurable preprocessing pipelines, portable YAML configuration files, standardized dataset wrappers, and an integrated collection of baseline forecasting methods. Together, these components reduce the engineering effort required to reproduce experiments, establish strong baselines, and systematically compare forecasting approaches under identical preprocessing and evaluation settings, while allowing the framework to be readily extended with additional datasets and models.

Another important contribution of \Tool{} is the promotion of comprehensive model evaluation. In addition to conventional predictive metrics, the framework incorporates clinically relevant evaluation measures and model complexity indicators, including the number of trainable parameters and FLOPs. This multidimensional benchmarking strategy enables researchers to analyze the trade-offs between predictive performance, clinical relevance, and computational complexity. The benchmarking results further illustrate that improvements in predictive performance are not necessarily accompanied by proportional increases in computational complexity, emphasizing the importance of jointly considering accuracy and efficiency when selecting forecasting models.

The usability study complements the technical evaluation by demonstrating that the proposed framework is accessible in practice. Participants reported a mean SUS score well above the commonly accepted benchmark, indicating that the graphical interface effectively supports reproducible experimentation for users with different levels of programming expertise. These results suggest that standardized experimental workflows can be made accessible without sacrificing flexibility.

\section{Conclusion}\label{conclusion}
In this work, we presented \Tool{}, a unified and extensible framework for reproducible experimentation with blood glucose time-series data. This standardization enables more reliable comparison of forecasting methods and provides a solid foundation for future methodological developments in blood glucose prediction research.

Our experimental evaluation highlights the importance of standardized preprocessing and evaluation protocols for blood glucose prediction research. By combining configurable preprocessing pipelines, standardized dataset wrappers, a comprehensive collection of baseline forecasting methods, integrated benchmarking, and an intuitive graphical user interface, \Tool{} enables transparent, reproducible, and comparable evaluation of forecasting approaches. The usability study further demonstrates that these capabilities are accessible to users with different levels of programming expertise, supporting the adoption of reproducible experimental workflows across a broad research community.

The modular architecture of \Tool{} is designed to accommodate future methodological advances. Additional datasets, preprocessing strategies, forecasting models, and evaluation metrics can be integrated through standardized interfaces while preserving compatibility with existing workflows. Future work will focus on expanding the collection of supported datasets and forecasting methods, incorporating additional preprocessing and evaluation techniques, further enhancing the framework's visualization and analysis capabilities, and broadening the applicability of \Tool{} beyond Type 1 Diabetes to other diabetes management scenarios, including Type 2 Diabetes.

\Tool{} addresses a key need in blood glucose prediction research by providing a standardized infrastructure for rigorous experimentation. Rather than introducing another forecasting architecture, it establishes a unified framework for the rigorous and systematic evaluation of both existing and future prediction methods. This standardization enables more reliable comparison of forecasting methods and provides a solid foundation for future methodological developments in blood glucose prediction research.

\section*{Acknowledgment}
This work was funded by the National Plan for NRRP Complementary Investments (PNC, established with the decree-law 6 May 2021, n. 59, converted by law n. 101 of 2021) in the call for the funding of research initiatives for technologies and innovative trajectories in the health and care sectors (Directorial Decree n. 931 of 06-06-2022) - project n. PNC0000003 - AdvaNced Technologies for Human-centrEd Medicine (project acronym: ANTHEM)~\footnote{\url{https://fondazioneanthem.it/}}. This work reflects only the authors’ views and opinions, neither the Ministry for University and Research nor the European Commission can be considered responsible for them.

Computational resources provided by hpc-ReGAInS@DISCo (\url{https://www.disco.unimib.it}).

\section*{References}
\bibliographystyle{IEEEtran}
\bibliography{bibliography}

\end{document}